\documentclass[letterpaper, 10 pt, conference]{ieeeconf}  

\IEEEoverridecommandlockouts                              

\usepackage{amsmath}
\usepackage{graphics}
\usepackage{graphicx}
\usepackage{amssymb}
\usepackage{color}
\usepackage{upgreek}
\usepackage{bbm, bbold}
\usepackage{algpseudocode}
\usepackage{lipsum}

\usepackage{tensor}
\usepackage[ruled,vlined]{algorithm2e}

\usepackage[normalem]{ulem}
\newcommand{\stkout}[1]{\ifmmode\text{\sout{\ensuremath{#1}}}\else\sout{#1}\fi}

\usepackage{textcomp} 

\usepackage{multicol}
\usepackage{multirow}
\usepackage{booktabs}

\usepackage[style=ieee]{biblatex}

\DeclareUnicodeCharacter{0301}{\'{e}} 

\DeclareSourcemap{
  \maps{
    \map{
      \pertype{article}
      \step[fieldset=language, null]
      \step[fieldset=url, null]
      \step[fieldset=doi, null]
      \step[fieldset=issn, null]
      \step[fieldset=isbn, null]
      \step[fieldset=note, null]
      \step[fieldset=editor, null]
      \step[fieldset=urldate, null]
      \step[fieldset=file, null]
    }
  }
}
\DeclareSourcemap{
  \maps{
    \map{
      \pertype{inproceedings}
      \step[fieldset=language, null]
      \step[fieldset=url, null]
      \step[fieldset=doi, null]
      \step[fieldset=issn, null]
      \step[fieldset=isbn, null]
      \step[fieldset=note, null]
      \step[fieldset=editor, null]
      \step[fieldset=urldate, null]
      \step[fieldset=file, null]
    }
  }
}
\DeclareSourcemap{
  \maps{
    \map{
      \pertype{incollection}
      \step[fieldset=language, null]
      \step[fieldset=url, null]
      \step[fieldset=doi, null]
      \step[fieldset=issn, null]
      \step[fieldset=isbn, null]
      \step[fieldset=note, null]
      \step[fieldset=editor, null]
      \step[fieldset=urldate, null]
      \step[fieldset=file, null]
    }
  }
}

\begin{document}

\title{\LARGE \bf  \textit{Aerobat}, A Bioinspired Drone to Test High-DOF Actuation and Embodied Aerial Locomotion}

\author{Alireza Ramezani$^{1\,*}$ and Eric Sihite$^{2}$%
\thanks{$^{1}$ The author is with the SiliconSynapse Laboratory, Department of Electrical and Computer Engineering, Northeastern University, Boston, MA-02119,USA.
    {Email: \tt\small a.ramezani@northeastern.edu}}%
\thanks{$^{2}$ The authors are with the Graduate Aerospace Laboratories of the California Institute of Technology (Caltech GALCIT), Pasadena, CA-91125, USA.
        {Emails: \tt\small esihite@caltech.edu}}%
\thanks{$^{*}$ This author is the corresponding author.}%
}



%

\maketitle

\begin{abstract}
This work presents an actuation framework for a bioinspired flapping drone called \textit{Aerobat}. This drone, capable of producing dynamically versatile wing conformations, possesses 14 body joints and is tail-less. Therefore, in our robot, unlike mainstream flapping wing designs that are open-loop stable and have no pronounced morphing characteristics, the actuation, and closed-loop feedback design can pose significant challenges. We propose a framework based on integrating mechanical intelligence and control. In this design framework, small adjustments led by several tiny low-power actuators called primers can yield significant flight control roles owing to the robot's computational structures. Since they are incredibly lightweight, the system can host the primers in large numbers. In this work, we aim to show the feasibility of joints' motion regulation in Aerobat's untethered flights.

\end{abstract}

\IEEEpeerreviewmaketitle

\definecolor{green}{rgb}{0.96, 0.29, 0.54}

\section{Introduction}

Bats' dynamic morphing wings are known to be extremely high-dimensional, involving the synchronous movements of many active and passive coordinates, joint clusters, in a gaitcycle. These animals apply their unique array of specializations to dynamically morph the shape of their wings to enhance their agility and energy efficiency. Copying bat dynamic morphing wing can bring fresh perspectives to micro aerial vehicle (MAV) design.

For instance, bats employ the combination of inertial dynamics and aerodynamics manipulations to showcase extremely agile maneuvers. Unlike rotary- and fixed-wing systems wherein aerodynamic surfaces (e.g., ailerons, rudders, propellers, etc.) come with the sole role of aerodynamic force adjustments, the articulated wings in bats possess more sophisticated roles \cite{riskin_bats_2009}. Or, it is known that bats can perform zero-angular-momentum turns by making differential adjustments (e.g., collapsing armwings) in the inertial forces led by their wings. Bats can apply a similar mechanism to perform sharp banking turns \cite{riskin_upstroke_2012,iriarte-diaz_whole-body_2011}. 

However, unfortunately, copying bat dynamic morphing wing flight is a significant ordeal. Existing bioinspired MAV designs completely overlook bat dynamic morphing capabilities because of the challenges associated with hardware design and control. As a result, much attention has been paid to simpler forms of animal aerial locomotion, such as those from insects. While the mathematical models of insect-inspired robots of varying size and complexity are relatively well developed, models of airborne, fluidic-based vertebrate locomotion, their control, and high-dimensional actuation remain largely open to date. 

\begin{figure}[t]
\vspace{0.1in}
    \centering
    \includegraphics[width=1.0\linewidth]{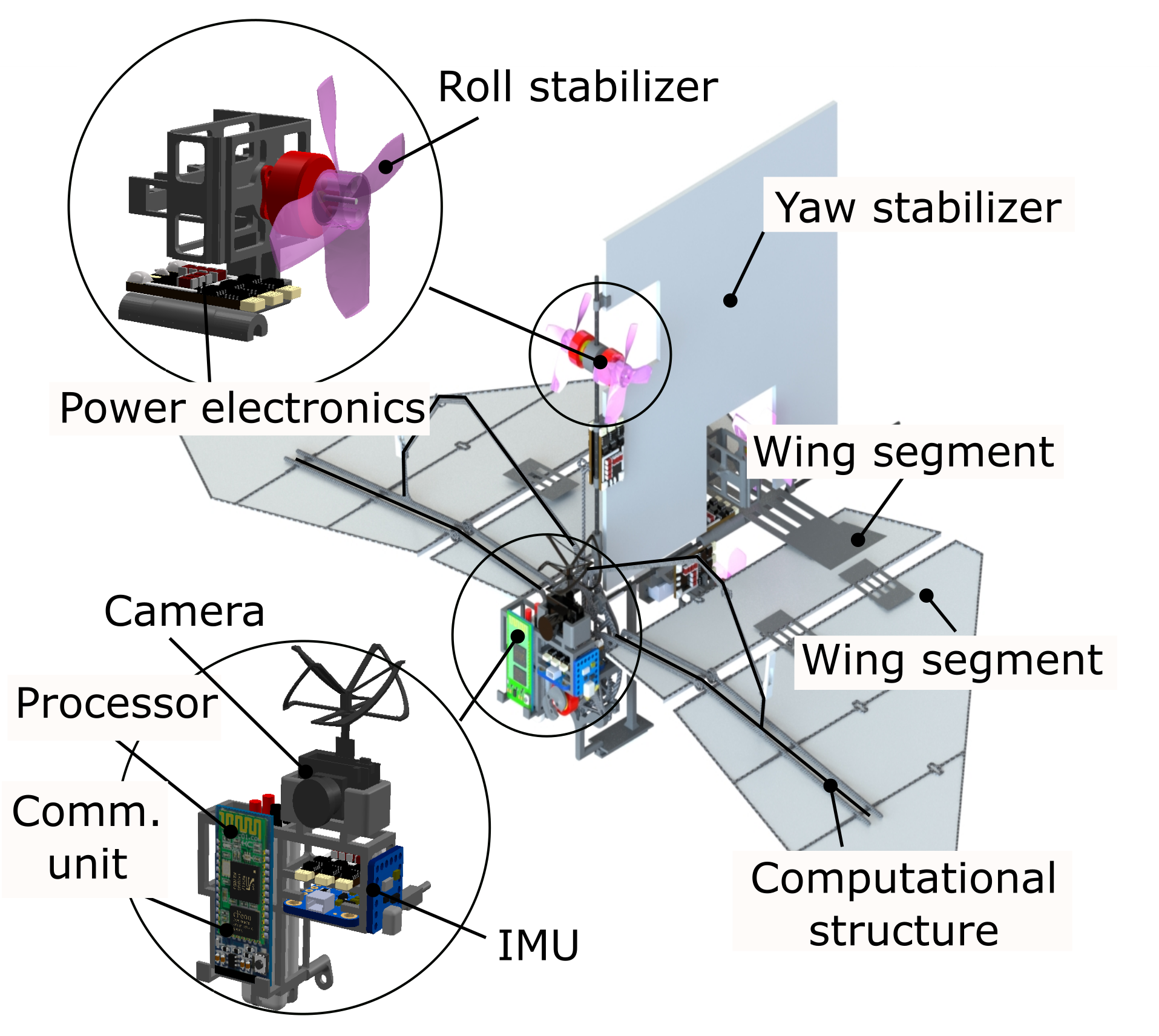}
    \caption{CAD depiction of our bioinspired robotic platform, Aerobat, augmented with small propellers for stabilization purpose.}
    \label{fig:cover}
\vspace{-0.1in}
\end{figure}

The mainstream school of thought inspired by insect flight has conceptualized wing as a mass-less, rigid structure, which is nearly planar and translates -- as a whole or in two-three rigid parts -- through space \cite{farrell_helbling_review_2018,zhang_instantaneous_2016,karpelson_review_2008}. In this view, wings possess no inertial effect, yield two-time-scale dynamics \cite{bullo_control_1995}, permit quasi-static external force descriptions \cite{lentink_biofluiddynamic_2009}, and produce a tractable dynamical system. Unfortunately, these paradigms fail to provide insight into airborne, vertebrate locomotion and lack the ingredients of a more complete and biologically meaningful model.

So, the overarching objective of our efforts is to present a systematic method for high-dimensional actuation in MAVs based on computational structure design and the optimal placement of low-power actuators, \textit{'primers'}, within computational structures \cite{sihite_computational_2020, sihite_enforcing_2020, sihite_integrated_2021, sihite2022unsteady, sihite2021orientation, lessieur2021mechanical, ramezani_biomimetic_2017, ramezani_bat_2016, ramezani_modeling_2016, ramezani_describing_2017}. Note that computational structures (also called mechanical intelligence or computational morphology) are mechanical structures that deliver computational resources. Therefore, the main contribution of this work is to demonstrate the feasibility of embodied aerial locomotion through simulation and experiment. Embodied locomotion is a notion that despite its endorsement by legged locomotion it has remained unexplored in morphing MAV design.  

We are designing morphing MAVs that capture bat dynamic morphing capabilities. Our designs so far have been tested in our lab. However, we have modified our bioinspired morphing MAVs such that flight stabilization is possible in outdoor through closed-loop feedback control of air jet. The resulting platform, shown in Fig.~\ref{fig:cover}, has allowed us to test dynamic morphing wing flight in realistic flight scenarios comparable to bat flights.  

Because of prohibitive design restrictions such as limited payload and power budget the application of classical joint motion control based on sensing, processing, and actuation is infeasible in this morphing MAV. The framework that we propose in this work has allowed the fast activation and regulation of many actuated degrees-of-freedom (DOF) in untethered flights. Our objective in this paper is to inspect the feasibility of gait regulation in Aerobat in untethered flights. For now, the air jet stabilizers are utilized for flight stabilization.

\section{Actuation Challenges}

The Aerobat shown in Fig.~\ref{fig:cover} has the following properties: (i) each wing is multi-segment and its mass constitutes a considerable part of the total weight, (ii) each wing is articulated, (iii) and its joints require high torque at large angular speeds. In the face of the actuation challenges in this MAV, the above properties call for a new form of computational resources different from that offered by classical closed-loop feedback. Specifically speaking these challenges include (i) attaining fast, high-dimensional, and synchronized joint motion (Challenge \# 1: Gait Generation) and (ii) achieving differential joint motion (Challenge \# 2: Gait Regulation) in the face of two limitations that follows.

In our MAV, small volume (space) dictates the allowable total power for mechanical work, which yield two issues. First, the robot can accommodate only a small number of powerful actuators. Second, the time an actuator requires to respond to its input is much larger than the gaitcycle time periods expected from the dynamic morphing MAV (i.e., one-tenth of a second) which makes within-gait joint motion regulation impossible.

\section{Use of Computational Morphology to Mitigate Actuation Challenges}

The flight dynamics of the morphing MAV can be described with a nonlinear system of the following form:
\begin{equation}
\begin{aligned}
\Sigma_{Full}&:\left\{
\begin{aligned}
    \dot x &= f(x) + g_1(x) u + g_2(x) y_2 \\
    y_1 &= h_1(x)\\
\end{aligned}
\right.\\
\Sigma_{Aero}&:\left\{
\begin{aligned}
    \dot \xi &= A_\xi (t) \xi + B_\xi (t) y_1 \\
    y_2 &= C_\xi (t)\xi + D_\xi (t) y_1
\end{aligned}
\right.
\end{aligned}
\label{eq:ss-rep-fulldyn}
\end{equation}
\noindent where $t$, $x$, and $\xi$ denote time, the state vector and aerodynamic hidden variables, respectively. In Eq.~\ref{eq:ss-rep-fulldyn}, the state vector $x$ embodies the position and velocity of the active $q_a$ and passive $q_p$ coordinates. The nonlinear terms given by $f(x)$ and $g_1(x)$ are obtained from Lagrange equations and embody inertial, Coriolis and gravity terms. The state-dependent matrices $g_{1}(x)$ and $g_{2}(x)$ map the joint actions $u(x)$ and external force $y_2(x)$ to the state velocity vector $\dot{x}$, respectively.

\begin{figure}[t]
\vspace{0.1in}
    \centering
    \includegraphics[width=0.9\linewidth]{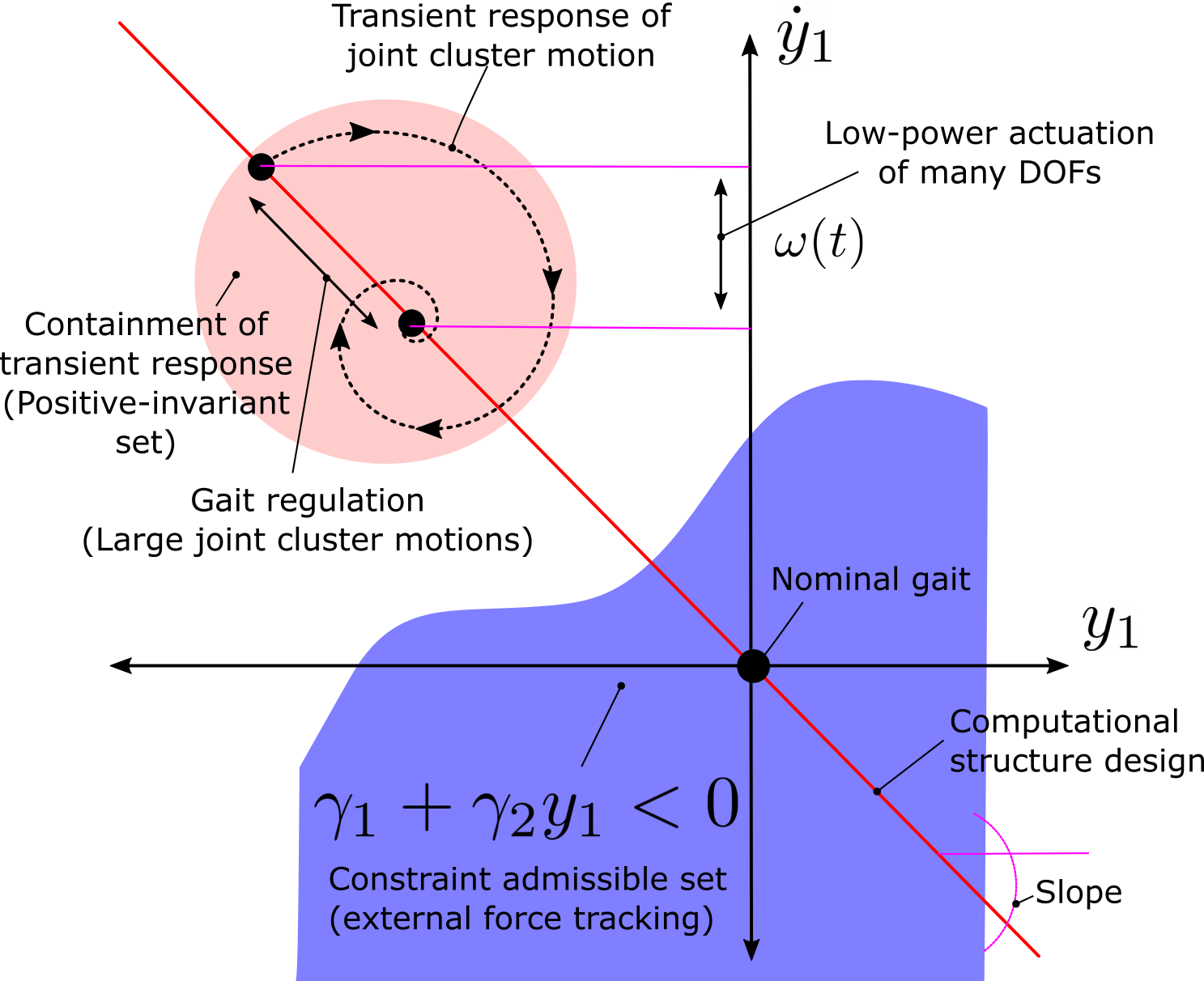}
    \caption{Illustrates the locus of equilibrium points from the Reference Governor (RG) model Eq.~\ref{eq:RG-model-simplified}. Note that the variable $\omega(t)$ can be leveraged to push the response inside the constraint-admissible set (dark blue region) for force tracking ($y_2 = C_\xi (t)\xi + D_\xi (t) y_1$ from Eq.~\ref{eq:ss-rep-fulldyn} forms the constraint-admissible set).}
    \label{fig:qs-ss-resp-adjust}
\vspace{-0.1in}
\end{figure}

The aerodynamic force output denoted by $y_2$ gives the instantaneous external forces. The governing dynamics are given by the state-space form made of $A_\xi$, $B_\xi$, $C_\xi$, and $D_\xi$ matrices \cite{izraelevitz_state-space_2017, boutet_unsteady_2018}
These terms are obtained based on Wagner indicial model and Prandtl lifting line theory reported in fluid dynamics textbook. The benefit of this indicial model is that it allows to efficiently compute the wake structures based on horseshoe vortex shedding. As a result, it allows to use wake structures to describe locomotion gaits. Wake-structure-based gaits are widely used in biology to describe bat aerial locomotion \cite{hubel_wake_2010}.

Gait generation and regulation are two main contributions of control inputs in Eq.~\ref{eq:ss-rep-fulldyn}. Dynamic morphing enforces tight requirements (such as power density and curse of dimensionality) on the input vector. By considering the holonomic constraint $y_1=h_1(x)$ we leave room for ourselves to be able to dichotomize the contributions from input $u$. Meaning, we can systematically determine which actuator generates and which one regulates the gait. 

So far there has been no clear strategy in the literature in that how these contributions can be systematically assigned to the actuators in a locomotion system. This view is majorly inspired by bats. In their membraned flight apparatus, bats possess specialized power and steering muscles that can generate and regulate gaits. Joint motion control by assuming a similar role for all actuators has been widely utilized in large systems such as manipulators and legged systems which possess less prohibitive design restrictions.

\begin{figure}[t]
\vspace{0.1in}
    \centering
    \includegraphics[width=0.8\linewidth]{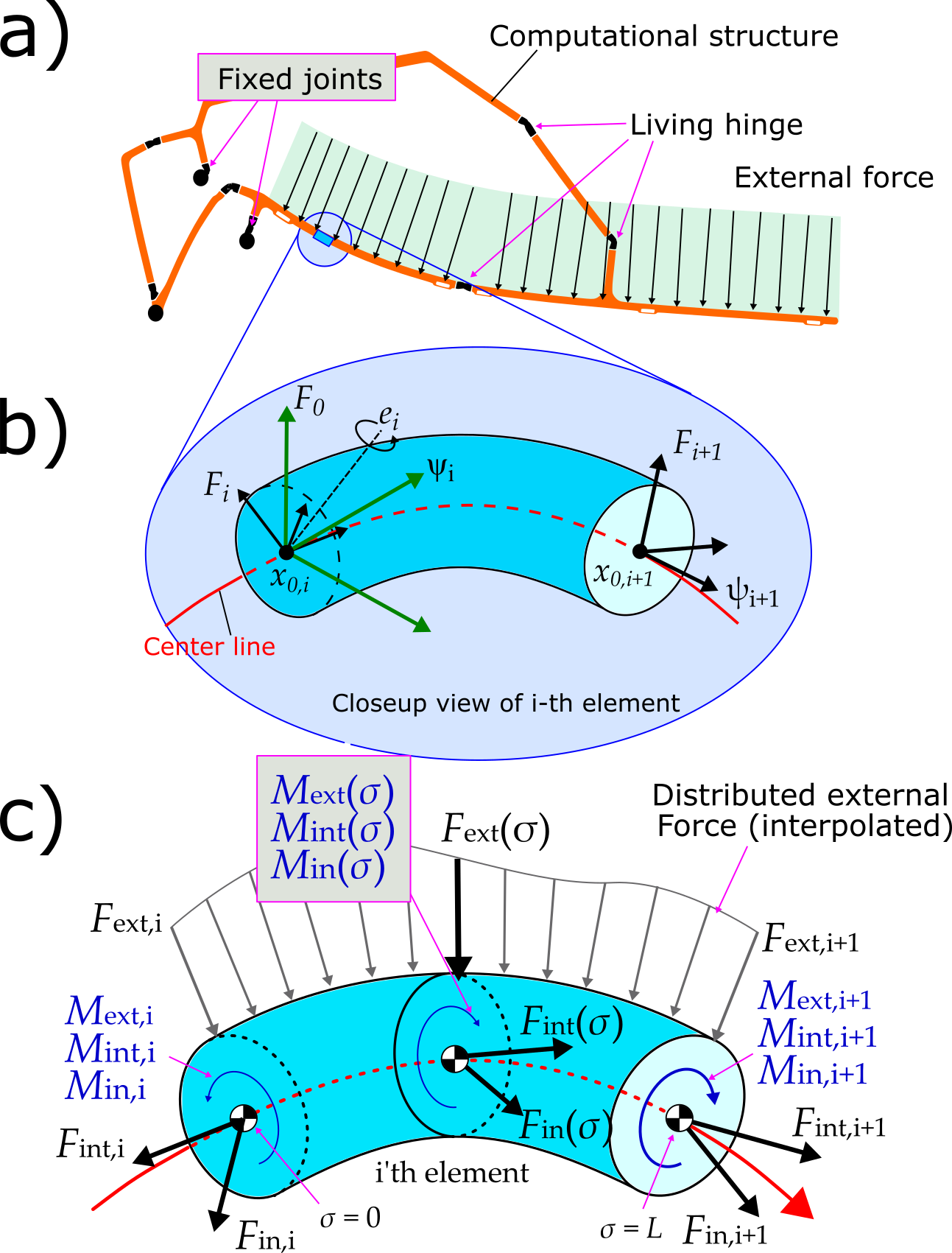}
    \caption{Cartoon depiction of a discretized computational structure.}
    \label{fig:actuator-placement}
\vspace{-0.1in}
\end{figure}

Also note that Eq.~\ref{eq:ss-rep-fulldyn} poses challenging control issues too. For instance, what is strikingly different from the flight dynamics of an insect-style system extensively reported in literature is that the two constraints, i.e., $y_1(x)$ and $y_2(x)$, have to be concurrently satisfied. We refer to this as two-fold tracking problem, known to biologists, however not endorsed by MAV engineers. Note that enforcing $y_1(x)$ and $y_2(x)$ yields dynamic morphing and desired force tracking, respectively. These two tracking problems are both important and cannot be compromised, therefore, they have to be enforced concurrently. 

Next, we show how assuming the rigid dichotomy in control contributions, i.e., gait generation and regulation, by assuming $y_1=h(x)$ can help solve the actuation and control challenges mentioned above. Let us motivate our solution first with a reference governor method which motivates the next steps in this paper. 

\section{Reference-Governor-Based View Towards Control of Eq.~\ref{eq:ss-rep-fulldyn}}

Briefly speaking, the objective is the generation and regulation of the joint cluster motion $q_a$ with a number of small actuators, primers. In general, these responsibilities are expected from $u$ however, our approach, is to subsume the input responsibilities under mechanical intelligence. The notion of reference governor (RG) can nicely fit into this context.

Specifically speaking, in Eq.~\ref{eq:ss-rep-fulldyn}, we can employ a new virtual input called primer $\omega(t)$ to regulate $y_2$ in a quasi-steady fashion. Consider the pre-stabilized version of Eq.~\ref{eq:ss-rep-fulldyn} given by
\begin{equation}
    \begin{aligned}
        \dot{x}_z &= f_z(x_z)+g_z(x_z)\omega(t)\\
    \end{aligned}
    \label{eq:RG-model-simplified}
\end{equation}
\noindent where $x_z=[q^\top_u,\dot q^\top_u]$. Here, it is assumed the dynamic state feedback $u$ is successfully designed and applied. As a result, $g_1(x)u$ in Eq.~\ref{eq:ss-rep-fulldyn} is re-written in a way that allows an affine-in-primer form given by Eq.~\ref{eq:RG-model-simplified}. 

Other state-dependent nonlinear terms are summarized under the nonlinear offset term $f_z(x_z)$ which denotes the restriction dynamics over the zero-dynamics manifold given by $\mathcal{Z}=\{x|y_1 = 0, L_f(y_1)=0\}$ where $L_f(.)$ is the Lie derivative. In Eq.~\ref{eq:RG-model-simplified}, the primer input governs the steady-state response of $y_1$ in the pre-stabilized system and can be designed for supervisory roles (control of external force $y_2$) with minimum actuation power.  

To see this better, consider the RG model given by Eq.~\ref{eq:RG-model-simplified} in the $y_1-\dot{y}_1$ space:  
\begin{equation} 
\left[\begin{array}{c}
\dot{y}_1 \\
\ddot{y}_1
\end{array}\right] = 
A_Y
\left[\begin{array}{c}
y_1\\
\dot{y}_1
\end{array}\right]
+
B_Y \omega(t)
\label{eq:output-dyn}
\end{equation}
\noindent where $A_Y$ and $B_Y$ are two matrices prescribed by the dynamic state feedback $u$. Notice that the role of the primer $\omega(t)$, which is comparable to the role of a disturbance term, is heavily determined by $A_Y$ and $B_Y$. These two matrices, which here are programmed into the system through feedback design in the RG model, can be systematically designed through robot design and actuator placement. 

The RG model concept is shown in Fig.~\ref{fig:qs-ss-resp-adjust}. As it can be seen the effectiveness of primer action in its ability to make large adjustments in the joint cluster motion $q_a$ is directly affected by the slope of the hyperplane shown in red. This slope is directly determined by $A_Y$ and $B_Y$ design. To see this, solve for the subset of state space that contains the steady-state solutions.

Next, we explain how the RG model can be realized in practice using the concept of embodied aerial locomotion.

\section{RG Model and Computational Structure Equivalency}

Embodied aerial locomotion emphasizes the fact that there must be an interconnection between the boundaries of morphology and control to regulate so many DOFs in morphing systems. For instance, bats have so many regulated DOFs that it is hard to believe they actively control all of them directly. 
 
Controllers lie in the space of abstract computation, and are usually implemented in computational layers or are programmed into the system. This trend can be widely seen in the world of larger robots (e.g., robot manipulators or legged systems). However, if robot morphology can also perform computation, it becomes possible for morphology to play a role of computation in the system which is useful for bioinspired MAVs. In other words, in the view of embodied aerial locomotion, part of the role of the controller is subsumed under computational morphology. 

\begin{figure}[t]
\vspace{0.1in}
    \centering
    \includegraphics[width=1.0\linewidth]{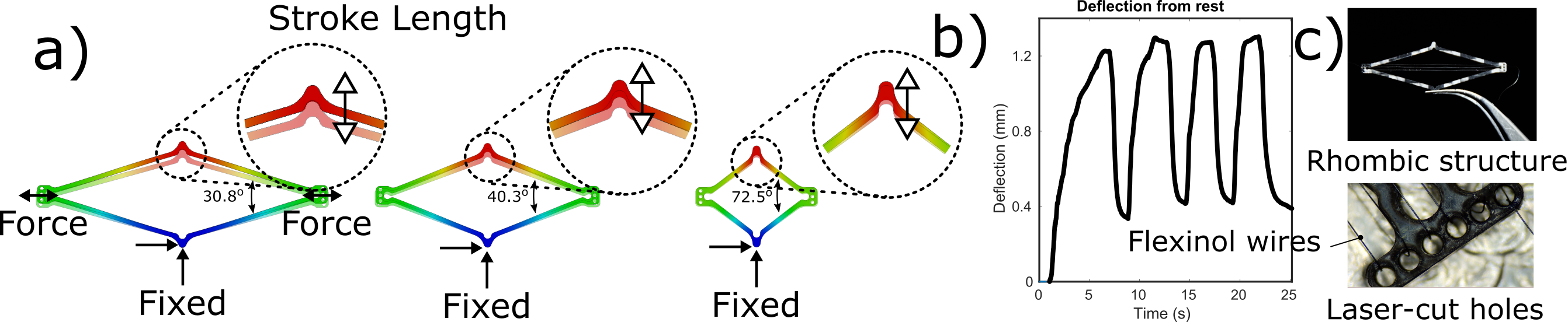}
    \caption{Shows a shape memory alloy (SMA)-based low-power actuator design (primer). A) Illustrates the laser-cut holes to recruit multiple loops of SMA wires to increase output force. B) Shows FEA analysis. C) A maxim 1.2 mm of displacement at the two ends of the rhombic structure can be achieved.}
    \label{fig:primer}
\vspace{-0.1in}
\end{figure}

Consider the cartoon depiction of the computational structure shown in Fig.~\ref{fig:actuator-placement}. This structure can be designed to deliver the response seen from $y_1=h(x)$ when closed-loop feedback is utilized. Meaning, the matrices $A_Y$ and $B_Y$ in Eq.~\ref{eq:output-dyn} can be hard-coded in the computational structure through structural (hardware) design in such a way that the stability ($A_Y$) and controllability ($A_Y$ and $B_Y$) properties of the RG model are captured. 

This approach gives us inexpensive computational resources. These resources are inexpensive in that they are needed anyways as part of the robot body structure. However, these structures' sole responsibility is not just hosting components but also contributing to the overall process of locomotion, a notion not endorsed in MAV design. To show the equivalency between the RG model and computational structures, we the constitute law widely used in smart structure analysis. The important point is that the proposed solution needs to be numerically inexpensive so that it can be utilized in the context of optimization.

\subsection{Constitute Law}
As shown in Fig.~\ref{fig:actuator-placement}, we discretize the computational structure. All elements are equally sized where $l$ denotes the element's length along the center line. The state vector for the i-th finite element, with the abuse of notation, is denoted by  $y_{1,i}=[x^\top_{0,i}, \Psi^\top_{i}, x^\top_{0,i+1}, \Psi^\top_{i+1}]^\top$. This state vector embodies the world position ($x_{0,i}$, $x_{0,i+1}$) and orientation ($\Psi_{i}$, $\Psi_{i+1}$) of the two cross-section ends in the i-th finite element. Now, the idea is to show that the output function response governed by Eq.~\ref{eq:output-dyn} can be reconstructed through mechanical design and low-power actuator placement (primer), namely, the realization given below is possible 

\begin{equation}
\begin{aligned}
    \begin{bmatrix}
    \dot y_{1,1}\\
    \ddot y_{1,1}\\
    \vdots\\
    \dot y_{1,n}\\
    \ddot y_{1,n}\\
    \end{bmatrix}=&
    \begin{bmatrix} 
    a_{11} & a_{12} & \dots \\
    \vdots & \ddots & \\
    a_{n1} &        & a_{nn} 
    \end{bmatrix}
    \begin{bmatrix}
    y_{1,1}\\
    \dot y_{1,1}\\
    \vdots\\
    y_{1,n}\\
    \dot y_{1,n}\\
    \end{bmatrix}
    +\\
    & \hspace{1cm} \begin{bmatrix} 
    b_{11} & b_{12} & \dots \\
    \vdots & \ddots & \\
    b_{n1} &        & b_{nn} 
    \end{bmatrix}
    \begin{bmatrix}
    \omega_1\\
    \vdots\\
    \omega_n\\
    \end{bmatrix}
\end{aligned}
    \label{eq:mimic_dynamics}
\end{equation}

\noindent where $y_{1,i}$ denotes the response from each element of the computational structure. By inspecting Eq.~\ref{eq:mimic_dynamics}, it can be seen that the input term $u$ contribution based on mode generation and regulation can be separately considered through the design of $a_{ij}$ (structure configuration and material properties) and $b_{ij}$ (low-power actuator placement). 

We take three steps to numerically obtain $a_{ij}$ and $b_{ij}$ as the function of structure configuration, material properties, and low-power actuator placement. Then, Eq.~\ref{eq:mimic_dynamics} is marched forward in time in Matlab for our analysis. Our analysis includes adjusting the material properties and physical properties of the computational structures to match the RG model.

\subsubsection{Interpolation Functions:} First, we define two scalar interpolation functions $N_{1,2}(\sigma)$, where the variable $\sigma\in[0,~l]$ denotes the distance traveled along the center line (see Fig.~\ref{fig:actuator-placement}). The interpolation functions vanish or take a fixed value at the boundaries, namely, $N_{1}(0)=0$, $N_{1}(l)=1$, $N_{2}(0)=1$, and $N_{2}(l)=0$. We use the interpolation functions to obtain the continuous forms for the discrete positions and orientations, that is, $x_0(\sigma)=\sum^2_{k=1} N_k(\sigma)x_{0,k}$ and $\Psi_0(\sigma)=\sum^2_{k=1} N_k(\sigma)\Psi_{0,k}$. 

\begin{figure}[t]
\vspace{0.1in}
    \centering
    \includegraphics[width=1.0\linewidth]{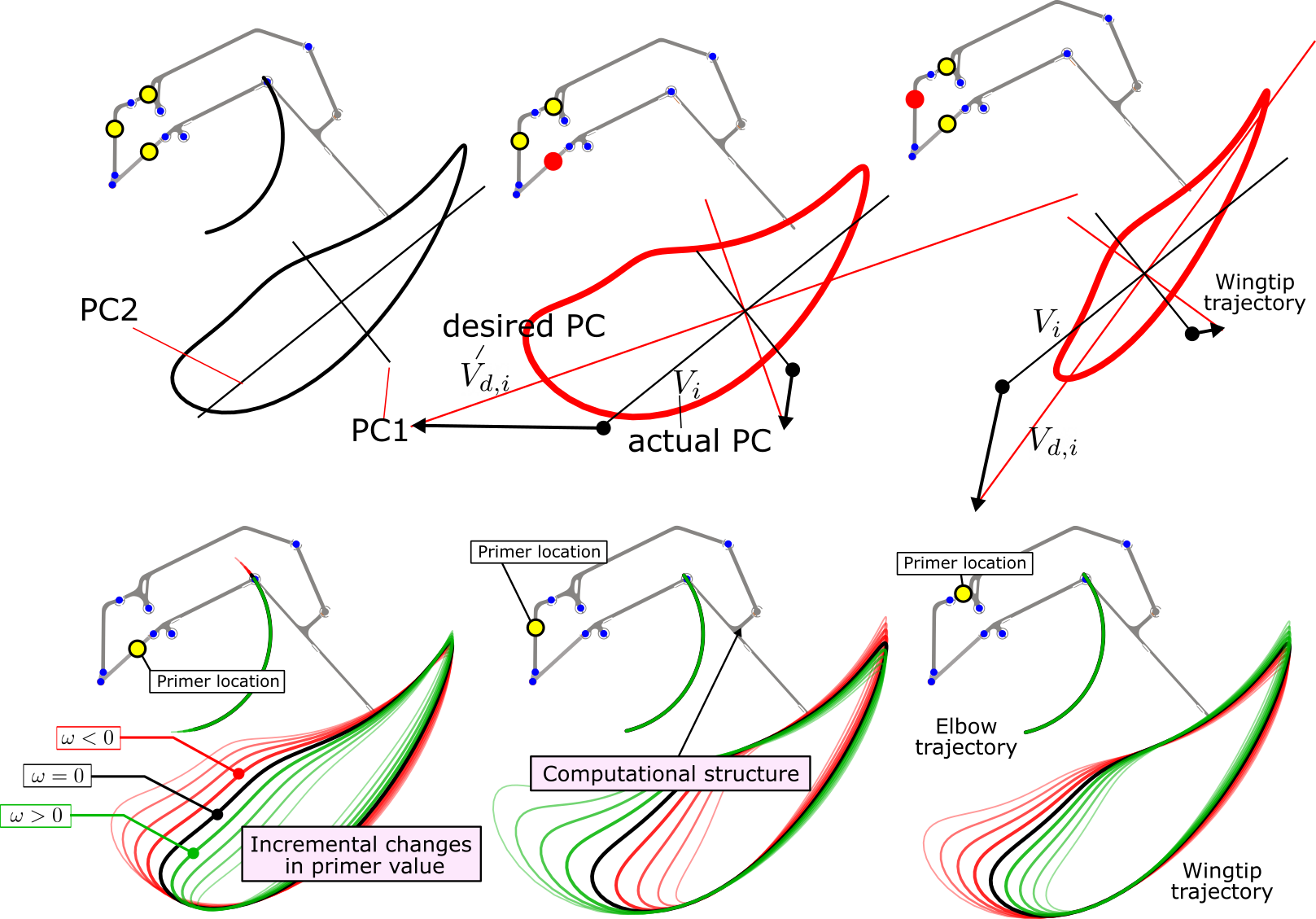}
    \caption{Top row illustrates optimal primer placement to achieve desired principle components (PC). $V_i$ and $V_{d,i}$ are the actual and desired PCs, respectively. Bottom row shows the response of the computational structure (front view) when optimally placed low-power primers are activated.}
    \label{fig:PCA}
\vspace{-0.1in}
\end{figure}

\subsubsection{Virtual Work:} Then, we obtain the increments in $\Delta \ddot y_{1,i}$, $\Delta \dot y_{1,i}$, and $\Delta y_{1,i}$ for each finite element following Hamilton's principle of virtual work. To do this, we consider three contributing forces and moments, including the internal {\small $F_1=[F^\top_{int}(\sigma)~M^\top_{int}(\sigma)]^\top$}, inertial {\small$F_2=[F^\top_{in}(\sigma)~M^\top_{in}(\sigma)]^\top$}, and external {\small$F_3=[F^\top_{ext}(\sigma)~M^\top_{ext}(\sigma)]^\top$} loads. We obtain the force increments by applying the first-order Taylor expansion to the element's static equilibrium defined by Hamiltion's principle, which yields $a_{ij}$ and $b_{ij}$ in Eq.~\ref{eq:mimic_dynamics}.

The variations (virtual displacements) of four terms are calculated in order to find the virtual works. The virtual displacement of the axial strain $\Gamma(\sigma)=R^\top x'_0$ (where $R$ is the rotation matrix and $x'_0=\frac{\partial x_0}{\partial \sigma}$ denotes the gradient of the world position $x_0$ with respect to $\sigma$) is obtained. The variation of the curvature $K(\sigma)=T(\Psi)\Psi'$ (where the tangent function is given by $T(\Psi)=I-\frac{1}{2!}\hat{\Psi}^2+\dots+\frac{(-1)^n}{(1+n)!}\hat{\Psi}^n$ and $\Psi'=\frac{\partial \Psi}{\partial \sigma}$) is obtained. Note that the wedge operator $\hat{.}$ transforms a vector to its skew symmetric matrix form. The infinitesimal displacement of the world position $x_0(\sigma)$ and rotation quasi-coordinates $\Theta(\sigma)$ of the cross-sections are obtained. 

\subsubsection{Incremental Load Changes:} The increments in the internal, inertial, and external loads form virtual mass-spring-damper dynamics for each element are given by 
\begin{equation}
\begin{aligned}
    \Delta F_1 =& \mathcal{I}(\kappa_1) \Delta y_{1,i}\\
    \Delta F_2 \, = &  \mathcal{I}(\kappa_2) \Delta \ddot y_{1,i} + 
  \mathcal{I}(\kappa_3) \Delta \dot y_{1,i} + 
  \mathcal{I}(\kappa_4) \Delta y_{1,i}\\
  \Delta F_3 = & \mathcal{I}(\kappa_5)\Delta y_{1,i} + \mathcal{I}(\kappa_6(\omega))\\
\end{aligned}
    \label{eq:delta_F}
\end{equation}
\noindent where $\mathcal{I}(\kappa_i)=\int^l_0 \kappa_i d\sigma$ and $\kappa_i$ are computed using the virtual displacements and interpolation functions. For example, $\kappa_{1}$ for internal and actuator loads are given by
\begin{equation}
\begin{aligned}
    \kappa_1 = &Q_2^\top Q_1^\top {\bf E} Q_1 Q_2\\
\end{aligned}
\end{equation}
\noindent where ${\bf E}$ is the diagonal matrix of material properties used in the computational structure and
\begin{equation}
\begin{aligned}
Q_1
&=
\begin{bmatrix}
    R^\top & 0 & \hat{\Gamma} T\\
    0 & T & \hat{K}T\Psi + T'
\end{bmatrix}\\
Q_2 &= \begin{bmatrix}
N'_1 & 0 & N'_2 & 0\\
0 & N'_1 & 0 & N'_2\\
0 & N_1 & 0 & N_2\\
\end{bmatrix}\\
\end{aligned}
\label{eq:variation_transformation_components}    
\end{equation}
\noindent Obtaining other $\kappa_i$ take a similar procedure that is omitted to save space. By stacking Eq.~\ref{eq:delta_F} for each element in the computational structure the $a_{ij}$ and $b_{ij}$ are obtained based on $\mathcal{I}(\kappa_i)$. Next, we show how Eq.~\ref{eq:mimic_dynamics} can be utilized for low-power actuator placement within Aerobat's computational structure to allow high-dimensional actuation.

\begin{figure}[t]
\vspace{0.1in}
    \centering
    \includegraphics[width=1.0\linewidth]{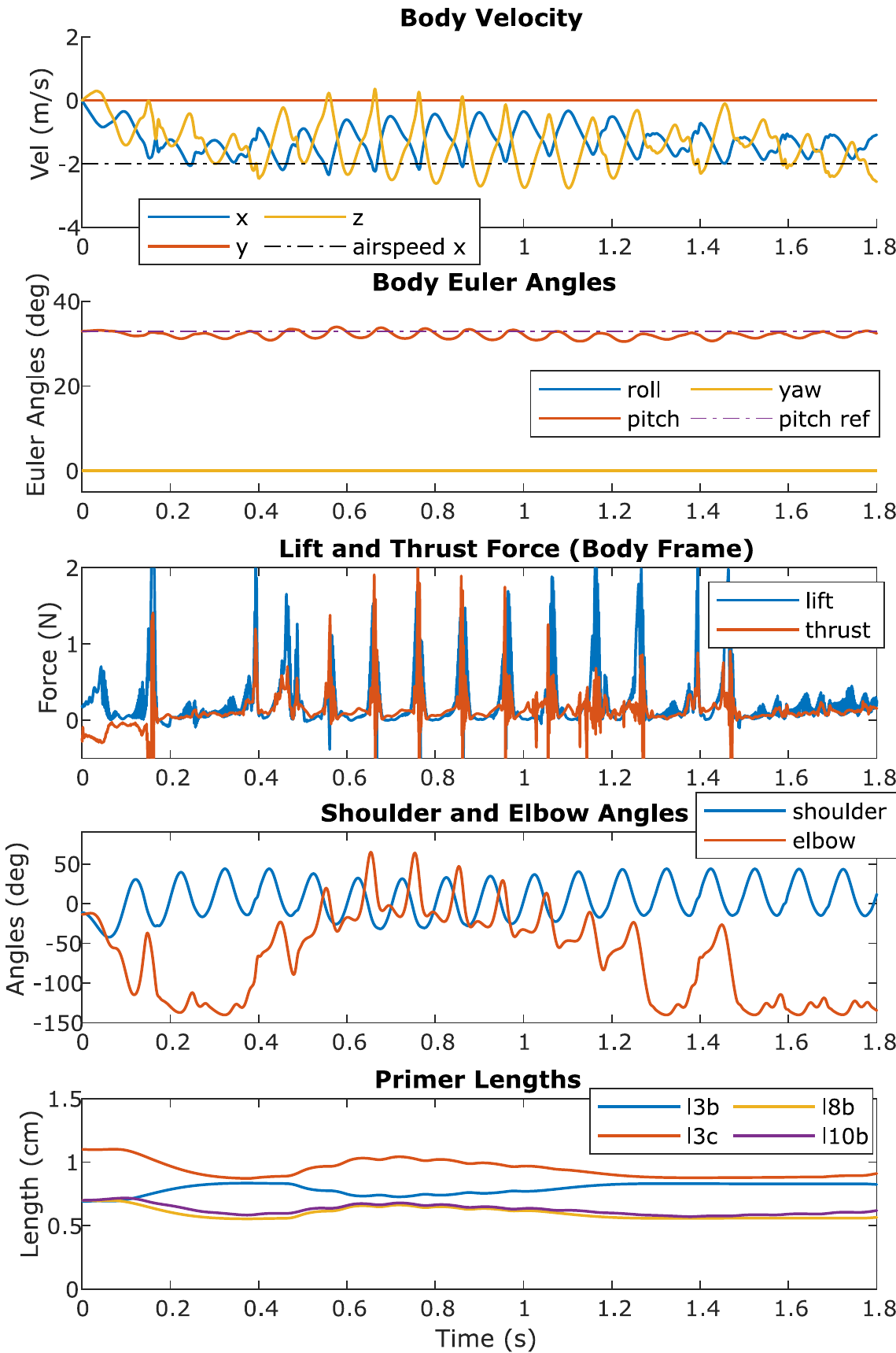}
    \caption{Simulated flight stabilization of Aerobat using primers' action. From top to bottom: Aerobat's main body velocity, orientation, generated external aerodynamic forces, wing joint trajectories (one wing), and primers' actions are illustrated.}
    \label{fig:simulation}
\vspace{-0.1in}
\end{figure}

\subsection{Optimal Low-Power Actuator (Primer) Placement}

Here, we obtain the optimal location for the low-power actuators (primers in Eq.~\ref{eq:RG-model-simplified}) within the computational structure by finding $b_{ij}$ in terms of $\kappa_i$. To find desired $B_Y$ matrix, we take the following approach. We consider the principle components (PCs) of the wingtip trajectory from the computational structure as shown in Fig.~\ref{fig:PCA}. These PCs are denoted by $V_i$. Then, we search for $B_Y$ matrices that rotate the PCs to the desired vectors $V_{d,i}$. This rotation should occur with a minimum actuation force from the primer. This actuator placement takes the following finite-state, nonlinear optimization form
\begin{equation}
\mathfrak{R}:
\left\{
\begin{aligned}
\underset{b_{ij}}{\textbf{\textrm{min}}} \quad & \omega^\top\omega\\
\textbf{\textrm{s.t.}} \quad & \dot Y - A_YY - B_Y\omega= 0\\
  & V_i-V_{d,i} = 0\\
  & \dot \xi -A_\xi \xi -B_\xi y_1 = 0\\
  & y_2 -C_\xi \xi -D_\xi y_1 = 0\\
\end{aligned}\right.
\label{eq:actuator_placement}
\end{equation}
\noindent In Eq.~\ref{eq:actuator_placement}, the last two equality constraints take into account the external aerodynamic forces from our model. Considering these external forces gives a more realistic prediction of the computational structure response and primer action under aeroelasticity.

\begin{figure*}[t]
\vspace{0.1in}
    \centering
    \includegraphics[width=1.0\linewidth]{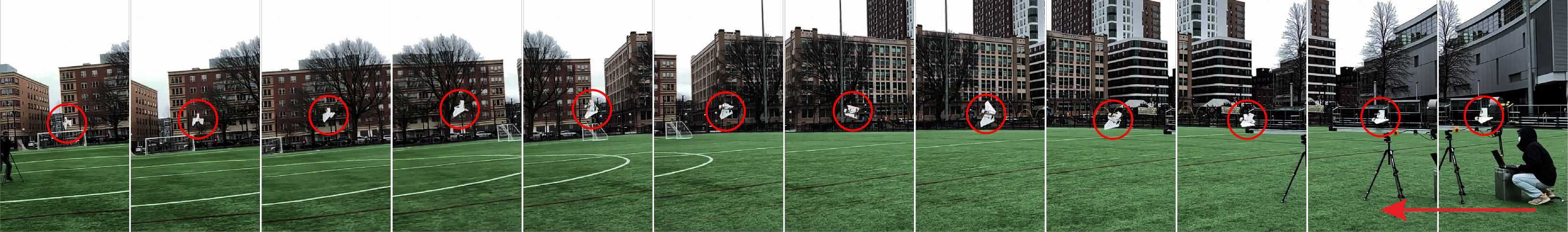}
    \caption{Shows snapshots of untethered flights stabilized with airjet (not primers). The untethered unit was utilized to verify the feasibility of high-dimensional actuation low-power actuators in small MAVs. There are prohibitive design restrictions in MAVs, e.g., limited payload and power budget which make high-dimensional actuation very challenging.}
    \label{fig:snapshot}
\vspace{-0.1in}
\end{figure*}

\section{Brief Overview of Primer Design}

We considered a few key properties in the design of primers. The primers must be extremely lightweight and easy to operate. We considered shape memory alloys (SMA) in the design of primers (see Fig.~\ref{fig:primer}). We found that SMA wires can provide large amount of forces in small form-factors and compliant structures, which can nicely fit into our design framework. Other than their compact designs, SMA ability to exert large contraction forces can be combined with the inherent compliance in the computational structure to create bi-directional actuators which need no power transmission mechanisms.

We used 38-$\mu$m Flexinol SMA wires which have actuation force of 20 gr. Their reaction time approximately is 0.25 sec. when supplied with 55 mA current. SMA wires generally shrink by approximately 4-5\% of their total length. Therefore, their stroke length must be amplified. In our next attempts, we used rhombus-shaped mechanical amplifiers (see Fig.~\ref{fig:primer}). In the rhombic primers, the SMA wires compress the two opposing edges of a rhombus-shaped structure. The result is amplified translational movements at the other opposing corners in the rhombus. These structures can be interesting for two reasons. First, their simple operation mechanisms and small footprints permit embedding them anywhere inside the structure with any output power and stroke length. The amount of force generated can simply be increased by having more loops of muscle wires. Second, they can be fabricated monolithically into the computational structure.

\section{Simulation and Experimental Results}
\label{sec:results}

We modeled the low-power actuators, primers, separately in SOLIDWORKS program and performed Finite Element Analysis (FEA) to estimate the stroke length and feasibility of different designs. The results are illustrated in Fig.\ref{fig:primer}. The FEA was done with the material properties of the interwoven carbon fiber plates used in the primers, i.e., Young's modulus of 40 GPa and density of 1600 kg/m$^3$. 

We evaluated the total displacement of the primer end-effector under the load given by the activated SMA wire. The actuation forces of the wires are estimated based on the datasheet \cite{flexinol_datasheet}, which is 20 gr for the 38-$\mu$m wire. This force is multiplied by the number of SMA loops in the design, which in our case is 11 loops for the 6 holes design. This results in a compression force of 440 gr.f on the actuated diagonal of the rhombic structure.

The FEA and experiments done on the 40mm-wide rhombus estimates a stroke length of a 1.04 mm. Placing this primer at the computational structures as shown in Fig.~\ref{fig:primer}, which has an elbow joint mean angular sensitivity of \mbox{-31$^\circ$/mm}, will greatly change the resulting wing trajectory (see Fig.~\ref{fig:PCA}). This significant change in the response of the structure was tested during untethered flights shown in Fig.~\ref{fig:snapshot} and Fig.~\ref{fig:PCA}. Note that in the untethered flight tests the primer action was not used to stabilize the flight. Instead we used air jet (see Fig.~\ref{fig:cover}) to actively stabilize roll and pitch dynamics. The untethered flight allowed us to check the feasibility of high-dimensional actuation in Aerobat.

In addition, based on the primer experimental results shown in Fig.~\ref{fig:primer}, the simulated closed-loop flight trajectories using Eq.~\ref{eq:ss-rep-fulldyn} were obtained. As shown in Fig.~\ref{fig:simulation}, the model can successfully maintain its target pitch reference by using its low-power actuators. Note that the tail-less system is pitch unstable. In these simulations, the primer length vary by up to approx. 1 mm. It can be seen that the mean elbow angle trajectory can shift by as much as 80$^\circ$. 

\section{Conclusion} 
\label{sec:conclusion}

This work presented an actuation framework based on integrating mechanical intelligence and low-power actuators (primers) for a bioinspired drone called Aerobat. Aerobat possesses 14 body joints, so actuating these joints can be very challenging. In addition, Aerobat is tail-less and open-loop unstable, which renders flight control extremely hard. In Aerobat, applying classical feedback design based on sensing, computation and actuation are not feasible because of: (i) a large number of degrees of freedom and (ii) prohibitive design restrictions such as limited payload and power budget. Therefore, the incorporation of state-of-the-art actuators in the system is not practical. The main objective of this paper is to show that the regulation of Aerobat joints in its untethered flight is feasible. We accomplished the aim by integrating small muscle-wire-based actuators within Aerobat's computational structure. Then, we used regulated air jets to permit untethered, stable flight. We attempted actuating Aerobat's joints using primers in untethered flights of Aerobat. 

\printbibliography

@article{zhang_instantaneous_2016,
	title = {Instantaneous wing kinematics tracking and force control of a high-frequency flapping wing insect {MAV}},
	volume = {11},
	language = {en},
	number = {1},
	urldate = {2020-12-05},
	journal = {Journal of Micro-Bio Robotics},
	author = {Zhang, Jian and Cheng, Bo and Deng, Xinyan},
	month = jun,
	year = {2016},
	pages = {67--84},
}

@article{riskin_bats_2009,
	title = {Bats go head-under-heels: the biomechanics of landing on a ceiling.},
	volume = {212},
	shorttitle = {Bats go head-under-heels},
	number = {Pt},
	journal = {The Journal of experimental biology},
	author = {Riskin, Daniel K. and Bahlman, Joseph W. and Hubel, Tatjana Y. and Ratcliffe, John M. and Kunz, Thomas Hans and Swartz, Sharon M.},
	year = {2009},
	pages = {945--953},
}

@article{lentink_biofluiddynamic_2009,
	title = {Biofluiddynamic scaling of flapping, spinning and translating fins and wings},
	volume = {212},
	language = {en},
	number = {16},
	urldate = {2020-03-02},
	journal = {Journal of Experimental Biology},
	author = {Lentink, D. and Dickinson, M. H.},
	month = aug,
	year = {2009},
	pages = {2691--2704},
}

@inproceedings{karpelson_review_2008,
	title = {A review of actuation and power electronics options for flapping-wing robotic insects},
	booktitle = {2008 {IEEE} {International} {Conference} on {Robotics} and {Automation}},
	author = {Karpelson, M. and {Gu-Yeon Wei} and Wood, R. J.},
	month = may,
	year = {2008},
	pages = {779--786},
}

@article{izraelevitz_state-space_2017,
	title = {State-{Space} {Adaptation} of {Unsteady} {Lifting} {Line} {Theory}: {Twisting}/{Flapping} {Wings} of {Finite} {Span}},
	volume = {55},
	shorttitle = {State-{Space} {Adaptation} of {Unsteady} {Lifting} {Line} {Theory}},
	language = {en},
	number = {4},
	urldate = {2021-01-18},
	journal = {AIAA Journal},
	author = {Izraelevitz, Jacob S. and Zhu, Qiang and Triantafyllou, Michael S.},
	month = apr,
	year = {2017},
	pages = {1279--1294},
}

@article{iriarte-diaz_whole-body_2011,
	title = {Whole-body kinematics of a fruit bat reveal the influence of wing inertia on body accelerations},
	volume = {214},
	language = {en},
	number = {9},
	urldate = {2019-09-04},
	journal = {Journal of Experimental Biology},
	author = {Iriarte-Diaz, J. and Riskin, D. K. and Willis, D. J. and Breuer, K. S. and Swartz, S. M.},
	month = may,
	year = {2011},
	pages = {1546--1553},
}

@article{farrell_helbling_review_2018,
	title = {A {Review} of {Propulsion}, {Power}, and {Control} {Architectures} for {Insect}-{Scale} {Flapping}-{Wing} {Vehicles}},
	volume = {70},
	language = {en},
	number = {1},
	urldate = {2020-05-22},
	journal = {Applied Mechanics Reviews},
	author = {Farrell Helbling, E. and Wood, Robert J.},
	month = jan,
	year = {2018},
}

@article{bullo_control_1995,
	series = {3rd {IFAC} {Symposium} on {Nonlinear} {Control} {Systems} {Design} 1995, {Tahoe} {City}, {CA}, {USA}, 25-28 {June} 1995},
	title = {Control on the {Sphere} and {Reduced} {Attitude} {Stabilization}},
	volume = {28},
	number = {14},
	urldate = {2019-09-10},
	journal = {IFAC Proceedings Volumes},
	author = {Bullo, F. and Murray, R. M. and Sarti, A.},
	month = jun,
	year = {1995},
	pages = {495--501},
}

@article{boutet_unsteady_2018,
	title = {Unsteady {Lifting} {Line} {Theory} {Using} the {Wagner} {Function} for the {Aerodynamic} and {Aeroelastic} {Modeling} of {3D} {Wings}},
	volume = {5},
	copyright = {http://creativecommons.org/licenses/by/3.0/},
	language = {en},
	number = {3},
	urldate = {2020-12-31},
	journal = {Aerospace},
	author = {Boutet, Johan and Dimitriadis, Grigorios},
	month = sep,
	year = {2018},
	pages = {92},
}

@article{riskin_upstroke_2012,
	title = {Upstroke wing flexion and the inertial cost of bat flight},
	volume = {279},
	language = {eng},
	number = {1740},
	journal = {Proceedings. Biological Sciences},
	author = {Riskin, Daniel K. and Bergou, Attila and Breuer, Kenneth S. and Swartz, Sharon M.},
	month = aug,
	year = {2012},
	pages = {2945--2950},
}

@article{ramezani_biomimetic_2017,
	title = {A biomimetic robotic platform to study flight specializations of bats},
	volume = {2},
	copyright = {Copyright © 2017, American Association for the Advancement of Science},
	issn = {2470-9476},
	url = {https://robotics.sciencemag.org/content/2/3/eaal2505},
	doi = {10.1126/scirobotics.aal2505},
	language = {en},
	number = {3},
	urldate = {2020-08-05},
	journal = {Science Robotics},
	author = {Ramezani, Alireza and Chung, Soon-Jo and Hutchinson, Seth},
	month = feb,
	year = {2017},
	note = {Publisher: Science Robotics
Section: Research Article},
}

@inproceedings{ramezani_modeling_2016,
	title = {Modeling and nonlinear flight controller synthesis of a bat-inspired micro aerial vehicle},
	url = {https://experts.illinois.edu/en/publications/modeling-and-nonlinear-flight-controller-synthesis-of-a-bat-inspi},
	language = {English (US)},
	urldate = {2020-05-09},
	booktitle = {{AIAA} {Guidance}, {Navigation}, and {Control} {Conference}},
	publisher = {American Institute of Aeronautics and Astronautics Inc, AIAA},
	author = {Ramezani, Alireza and Shi, Xichen and Chung, Soon-Jo and Hutchinson, Seth Andrew},
	month = jan,
	year = {2016},
}

@inproceedings{ramezani_describing_2017,
	address = {Cham},
	series = {Lecture {Notes} in {Computer} {Science}},
	title = {Describing {Robotic} {Bat} {Flight} with {Stable} {Periodic} {Orbits}},
	isbn = {978-3-319-63537-8},
	doi = {10.1007/978-3-319-63537-8_33},
	language = {en},
	booktitle = {Biomimetic and {Biohybrid} {Systems}},
	publisher = {Springer International Publishing},
	author = {Ramezani, Alireza and Ahmed, Syed Usman and Hoff, Jonathan and Chung, Soon-Jo and Hutchinson, Seth},
	editor = {Mangan, Michael and Cutkosky, Mark and Mura, Anna and Verschure, Paul F.M.J. and Prescott, Tony and Lepora, Nathan},
	year = {2017},
	pages = {394--405},
}

@inproceedings{ramezani_bat_2016,
	title = {Bat {Bot} ({B2}), a biologically inspired flying machine},
	doi = {10.1109/ICRA.2016.7487491},
	booktitle = {2016 {IEEE} {International} {Conference} on {Robotics} and {Automation} ({ICRA})},
	author = {Ramezani, Alireza and Shi, Xichen and Chung, Soon-Jo and Hutchinson, Seth},
	month = may,
	year = {2016},
	pages = {3219--3226},
}

@article{hubel_wake_2010,
	title = {Wake structure and wing kinematics: the flight of the lesser dog-faced fruit bat, {Cynopterus} brachyotis},
	volume = {213},
	issn = {0022-0949},
	shorttitle = {Wake structure and wing kinematics},
	url = {https://doi.org/10.1242/jeb.043257},
	doi = {10.1242/jeb.043257},
	number = {20},
	urldate = {2021-08-04},
	journal = {Journal of Experimental Biology},
	author = {Hubel, Tatjana Y. and Riskin, Daniel K. and Swartz, Sharon M. and Breuer, Kenneth S.},
	month = oct,
	year = {2010},
	pages = {3427--3440},
}

@article{sihite_computational_2020,
	title = {Computational {Structure} {Design} of a {Bio}-{Inspired} {Armwing} {Mechanism}},
	volume = {5},
	issn = {2377-3766, 2377-3774},
	url = {https://ieeexplore.ieee.org/document/9143405/},
	doi = {10.1109/LRA.2020.3010217},
	language = {en},
	number = {4},
	urldate = {2021-03-01},
	journal = {IEEE Robot. Autom. Lett.},
	author = {Sihite, Eric and Kelly, Peter and Ramezani, Alireza},
	month = oct,
	year = {2020},
	pages = {5929--5936}
}

@article{sihite_integrated_2021,
	title = {An {Integrated} {Mechanical} {Intelligence} and {Control} {Approach} {Towards} {Flight} {Control} of {Aerobat}},
	url = {https://arxiv.org/abs/2103.16566v1},
	language = {en},
	urldate = {2021-04-11},
	author = {Sihite, Eric and Darabi, Atefe and Dangol, Pravin and Lessieur, Andrew and Ramezani, Alireza},
	month = mar,
	year = {2021},
}

@inproceedings{sihite_enforcing_2020,
	address = {Jeju Island, Korea (South)},
	title = {Enforcing nonholonomic constraints in {Aerobat}, a roosting flapping wing model},
	isbn = {978-1-72817-447-1},
	url = {https://ieeexplore.ieee.org/document/9304158/},
	doi = {10.1109/CDC42340.2020.9304158},
	language = {en},
	urldate = {2021-03-01},
	booktitle = {2020 59th {IEEE} {Conference} on {Decision} and {Control} ({CDC})},
	publisher = {IEEE},
	author = {Sihite, Eric and Ramezani, Alireza},
	month = dec,
	year = {2020},
	pages = {5321--5327}
}

@misc{flexinol_datasheet,
    title={Technical Characteristics of FLEXINOL Actuator Wires},
    note={Available: \url{https://www.dynalloy.com/pdfs/TCF1140.pdf}},
    year={2017}
}

@article{sihite2022unsteady,
  title={Unsteady aerodynamic modeling of Aerobat using lifting line theory and Wagner's function},
  author={Sihite, Eric and Ghanem, Paul and Salagame, Adarsh and Ramezani, Alireza},
  journal={arXiv preprint arXiv:2207.12353},
  year={2022}
}

@inproceedings{sihite2021orientation,
  title={Orientation stabilization in a bioinspired bat-robot using integrated mechanical intelligence and control},
  author={Sihite, Eric and Lessieur, Andrew and Dangol, Pravin and Singhal, Akshath and Ramezani, Alireza},
  booktitle={Unmanned Systems Technology XXIII},
  volume={11758},
  pages={12--20},
  year={2021},
  organization={SPIE}
}

@inproceedings{lessieur2021mechanical,
  title={Mechanical design and fabrication of a kinetic sculpture with application to bioinspired drone design},
  author={Lessieur, Andrew and Sihite, Eric and Dangol, Pravin and Singhal, Akshath and Ramezani, Alireza},
  booktitle={Unmanned systems technology XXIII},
  volume={11758},
  pages={21--27},
  year={2021},
  organization={SPIE}
}

\end{document}